\documentclass{llncs}
\usepackage{graphicx}
\usepackage{graphics}
\usepackage{epsfig}
\usepackage{wrapfig}
\usepackage{amssymb,amsmath}
\usepackage{algpseudocode}
\usepackage[linesnumbered]{algorithm2e}
\usepackage{multirow}
\usepackage{url}

\usepackage[utf8]{inputenc} 
\usepackage[T1]{fontenc}    
\usepackage{hyperref}       
\usepackage{url}            
\usepackage{booktabs}       
\usepackage{amsfonts}       
\usepackage{nicefrac}       
\usepackage{microtype}      
\usepackage{amsmath}
\usepackage{graphicx}
\usepackage{multirow}		
\usepackage{color}
\usepackage{caption}		
\usepackage{array}
\newcolumntype{L}[1]{>{\raggedright\let\newline\\\arraybackslash\hspace{0pt}}m{#1}}
\newcolumntype{C}[1]{>{\centering\let\newline\\\arraybackslash\hspace{0pt}}m{#1}}
\newcolumntype{R}[1]{>{\raggedleft\let\newline\\\arraybackslash\hspace{0pt}}m{#1}}

\usepackage[textsize=scriptsize,backgroundcolor=yellow!40]{todonotes}
\usepackage{subcaption}

\begin{document}

\title{Online anomaly detection using statistical leverage for streaming business process events}	

\author{Jonghyeon Ko and Marco Comuzzi}

\institute{Department of Industrial Engineering\\ Ulsan National Institute of Science and Technology  (UNIST) \\ Ulsan, Republic of Korea \\
{\fontsize{8}{10}\selectfont \email{\{whd1gus2,mcomuzzi\}@unist.ac.kr}}}

\maketitle

\begin{abstract}
While several techniques for detecting trace-level anomalies in event logs in offline settings have appeared recently in the literature, such techniques are currently lacking for online settings. Event log anomaly detection in online settings can be crucial for discovering anomalies in process execution as soon as they occur and, consequently, allowing to promptly take early corrective actions. This paper describes a novel approach to event log anomaly detection on event streams that uses statistical leverage. Leverage has been used extensively in statistics to develop measures to identify outliers and it has been adapted in this paper to the specific scenario of event stream data. The proposed approach has been evaluated on both artificial and real event streams. 
\end{abstract}


\section{Introduction}

Information logged during the execution of business processes is available in so-called event logs, which contain events belonging to different process instances (or \emph{cases}). Each event is described by multiple attributes, such as a timestamp and a label capturing the activity in the process that was executed. 

Event logs are prone to errors, which can stem from a variety of root causes~\cite{AD2,AD9}, such as system malfunctioning or sub-optimal resource behaviour. For instance, sloppy human resources may forget to log the execution of specific activities in a process, or a system reboot may assign a different case id to all the new events recorded after rebooting. Errors in event log hamper the possibility of extracting useful process insights from event log analysis, and should therefore be fixed as early as possible~\cite{van2019online}. 

To this end, the research field of event log anomaly detection (or event log cleaning) has emerged recently, providing methods to detect anomalies at trace level~\cite{AD2,AD9,AD5,AD3,AD4}, i.e., concerning the order and occurrence of activities in a process, and at event level~\cite{AD7,AD10,AD11}, i.e., concerning the value of attributes of events, using a variety of different approaches. Note that event log anomaly detection is normally (process) model-agnostic, that is, it does not assume the existence of a process model or clean traces from which a model can be extracted. This aspect separates this research field from traditional process mining research on compliance checking~\cite{intro8}.

In the specific case of online settings, i.e., event streams, while research has recently emerged in the field of online compliance checking, only the work by Tavares et al.~\cite{tavares2019leveraging} addresses the issue of anomaly detection. In particular, the authors propose a method to detect point anomalies specified by Principal Component Analysis (PCA). These point anomalies, however, do not normally reflect  real-life anomaly patterns, such as inserting, skipping or switching events, commonly considered by event log anomaly detection in offline settings. Therefore, we argue that there is a lot of potential for new research in this area. 

More in general, event log anomaly detection in online settings can be crucial for discovering anomalies in process execution as soon as they occur and, consequently, allowing to promptly take early corrective actions. 
The online settings, however, obviously introduce additional challenges to the design of an event log anomaly detection method. In particular, owing to the finite memory assumption of online settings~\cite{burattin2017online,burattin2017framework,burattin2018online,tavares2019leveraging,van2019online}, only a limited number of (recent) events are available at any given time to take a decision. This prevents to apply effectively some of the approaches that have been proposed in the literature for event log anomaly detection in offline settings. Probabilistic methods that detect anomalies after having created an intermediate model of frequent process behaviour~\cite{AD2,AD3,AD4} are hampered by the fact that only a limited number of events may be available to create such models. Online settings also prevent the application of machine learning reconstructive techniques for anomaly detection, e.g.~\cite{AD10,AD14}. These, in fact, normally rely on deep learning models, which require a high number of data points (complete process traces in this scenario) to be trained effectively. Also, any update of these models may require a long training time.

In this paper we propose an information-theoretic approach to online event log anomaly detection at trace level. Specifically, we devise an anomaly score based on statistical leverage~\cite{leverage1}. The leverage is a relative measure of the information content of observations in a dataset that has been used extensively in statistics to develop observation distance measures and outlier detection techniques. Since leverage captures the information content of one observation in respect of all others in a dataset, the anomaly score proposed in this paper can always be calculated reliably based on the information available at any point in time, resulting in an anomaly detection method that does not require extensive amount of data to be executed effectively. 

After having presented the related work (in Section~\ref{sec:related works}),  Section~\ref{sec:methods} presents a trace anomaly score based on the notion of statistical leverage. Then, we discuss how this score can be applied to anomaly detection of streams of events, addressing issues such as the grace period, the finite memory assumption, and the identification of anomaly detection thresholds. The proposed method is evaluated (in Section~\ref{sec:eval}) on both artificial and real event logs injected with trace-level anomalies. Conclusions finally are drawn in Section~\ref{sec:conclu}.

\section{Related work}
\label{sec:related works}

While there is only limited literature regarding online event log anomaly detection, a number of recent contributions have focused on online conformance checking. To some extent, conformance checking can be seen as model-aware anomaly detection, since process models, given or extracted from clean traces, can be seen as signatures of positive behaviour to detect anomalies. 
As referred by~\cite{burattin2018online}, there are currently two research lines in online conformance checking: the prefix-alignment approach~\cite{van2019online} and the model-based approach~\cite{burattin2017online,burattin2017framework,burattin2018online}. 

Conformance checking/alignment of streaming events tends to overestimate the computation of optimal alignments. In order to avoid this issue, \cite{van2019online} provides the first incremental/online conformance checking technique that uses prefix-alignment. Prefix-alignment~\cite{van2019online} is characterised by high computational complexity and prevents to define a warming up period. Alternatively, Online Conformance Transition Systema (OCTS)~\cite{burattin2017online,burattin2017framework} can partially check compliance on regions of a process. This technique also suffers from high computational complexity and prevents to consider the warm start scenario. In \cite{burattin2018online}, the first solution to achieve a warm start with streaming events has been  proposed by introducing weak order relations, that have reduced computational complexity.

Regarding event log anomaly detection, as mentioned in the Introduction, Tavares et al.~\cite{tavares2019leveraging} have first  applied the online clustering algorithm DenStream~\cite{cao2006density} to detect anomalies on event streams. DenStream clusters cases into two groups, normal and anomalous, using histogram-based frequency of activities contained in each case. Since the histogram-based frequency ignores the sequence of events in traces, DenStream detects point anomalies in event logs defined by Principal Component Analysis (PCA)~\cite{leverage3}. 


\section{Research framework}
\label{sec:methods}

There are two different elements in the proposed framework: the anomaly score and the anomaly detection method. The former (presented in Section~\ref{sec:anomaly score}) concerns the definition of a trace anomaly score based on statistical leverage. The latter (Section~\ref{sec:strategy}) concerns setting a threshold value above which a trace is considered anomalous based on its anomaly score.

\subsection{Anomaly score}
\label{sec:anomaly score}

Statistical \textit{leverage}~\cite{leverage1} is a measure indicating  how far away each observation is scattered from other observations in a dataset. It has been used as a key support measure for developing different observation distance metrics,  such as Cook's distance, the Welsch-Kuh distance, and the Welsch's distance. 

Given a matrix $X$, with $X \in \mathbb{R}^{J \times I}$, of  a dataset with $J$ observations and $I$ numerical attributes (or variables), the leverage of the observations in $X$ are the diagonal elements of the projection matrix $H=X(X^{T}X)^{-1}X^{T}$. Specifically, the leverage of the $j$-th observation in $X$ is the diagonal element $h_{j,j} \in H$, which is comprised by definition between 0 and 1. The higher its leverage, the more likely an observation to be an anomaly.

Our objective is to detect anomalies at the level of occurrence and order of events in traces. Therefore, we can abstract an event log $E$ as a set of $J$ traces $\{ \sigma_j \}_{j=1,\ldots, J}$. Each trace is a sequence of events $e_{i,j}$ of variable length $N_j$, i.e., $\sigma_j = \{ e_{1,j}, \ldots, e_{N_j,j} \}$. Events are ordered in a trace by timestamp in ascending order and are defined by the activity that they represent, which is one in a set $A = \{ a_{1} \ldots, a_K \}$ of $K$ possible activity labels.


In order to define a leverage-based anomaly measure of traces in an event log, two pre-processing steps are necessary. The first one is an integer encoding step. This is necessary because the attributes of a dataset $X$ must be numerical to calculate $H$, while the activity attribute in event logs is categorical. Second, events in an event log must be aggregated at trace level, such that the resulting matrix $X$ has $J$ rows, i.e., one for each trace. In conclusion, the  projection matrix $H(E)$  can be calculated by considering an observation matrix  $X(E)$ obtained from $E$ applying the following pre-processing steps.

\begin{figure}[t]
    \centering
    \makebox[0pt]{%
    \includegraphics[width=0.4\paperwidth]{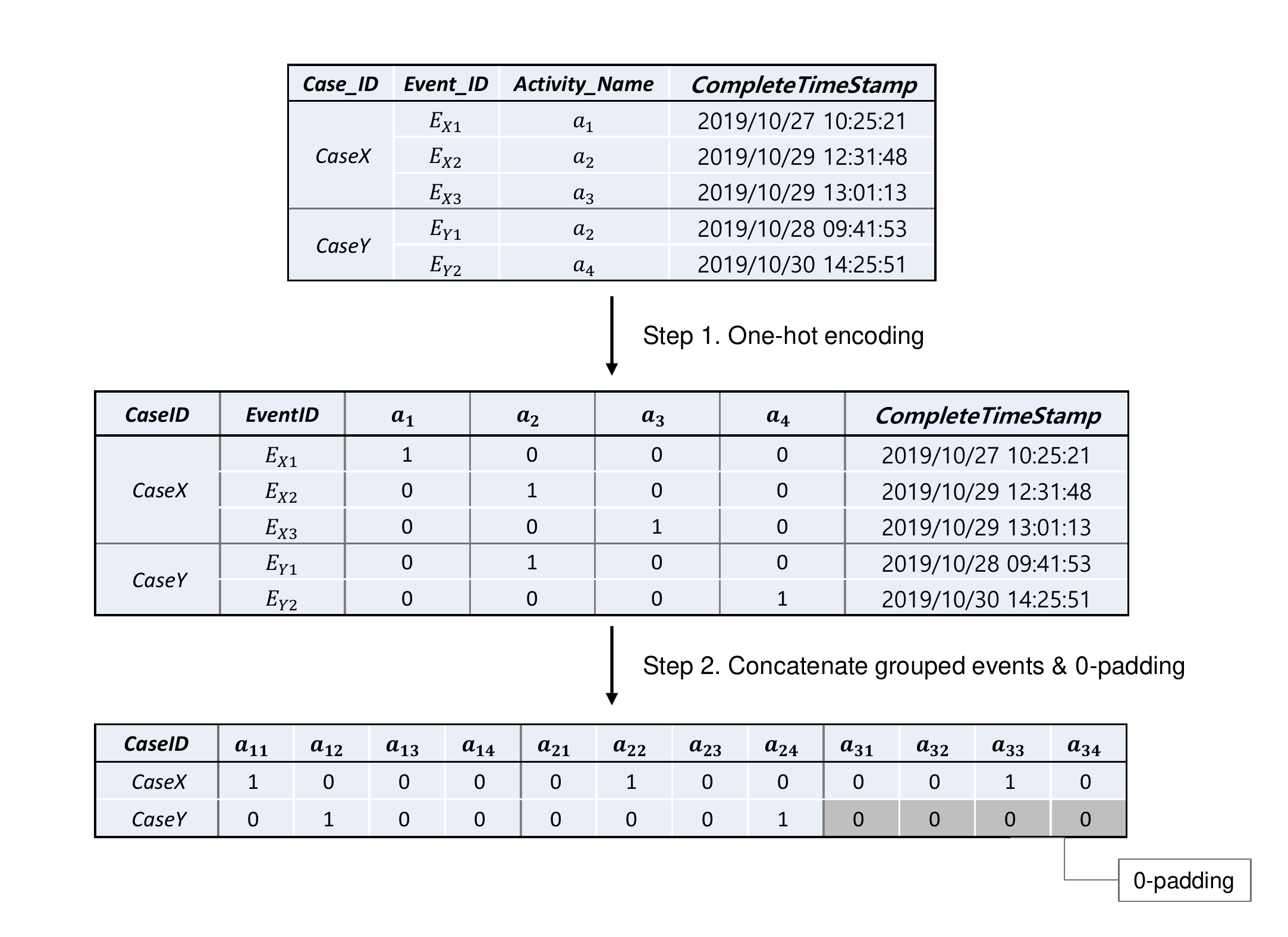}}
    \caption{One-hot encoding and 0-padding}
    \label{fig:preprocess}
\end{figure}

In the first pre-processing step (see Figure~\ref{fig:preprocess}), similarly to~\cite{AD10}, we apply one-hot encoding, that is, each event $e_{i,j}$ is encoded into a set  $K$ dummy attributes $d_{i,j,k}$ such that:

\[ d_{i,j,k} =
		\begin{cases}
		1       & \quad \text{if } e_{i,j} = a_k \\
		0  & \quad \text{otherwise}
		\end{cases}
\]

Then, for trace-level aggregation, the one-hot encoded events are horizontally concatenated for each trace. Since traces have different length, for the traces shorter  than the longest one(s) in $E$, i.e., with less events than $N^{max} = \max_{\sigma_j \in E}\{N_j\}$, zero padding is applied. For example, given a case consisting of 4 events and $N^{max} = 5$, the fifth event of this case is zero padded, therefore, $d_{5,j,k} = 0$, $\forall k$.  
Based on this pre-processing, an event log $E$ is encoded into an observation matrix $X(E)$ with $J$ rows (traces) and $I = N^{max} \times K$ columns (attributes).

Using  $X(E)$, we can now define a first leverage-based anomaly score $\hat{l}(\sigma_j)$ by extracting the diagonal elements of $H(E)=X(E)\cdot (X(E)^{T}\cdot X(E))^{-1}\cdot X(E)^{T}$: 

$$
\hat{l}(\sigma_j) = h_{j,j}
$$

\begin{figure}[th]
    \centering
    \makebox[0pt]{%
    \includegraphics[width=0.7\textwidth]{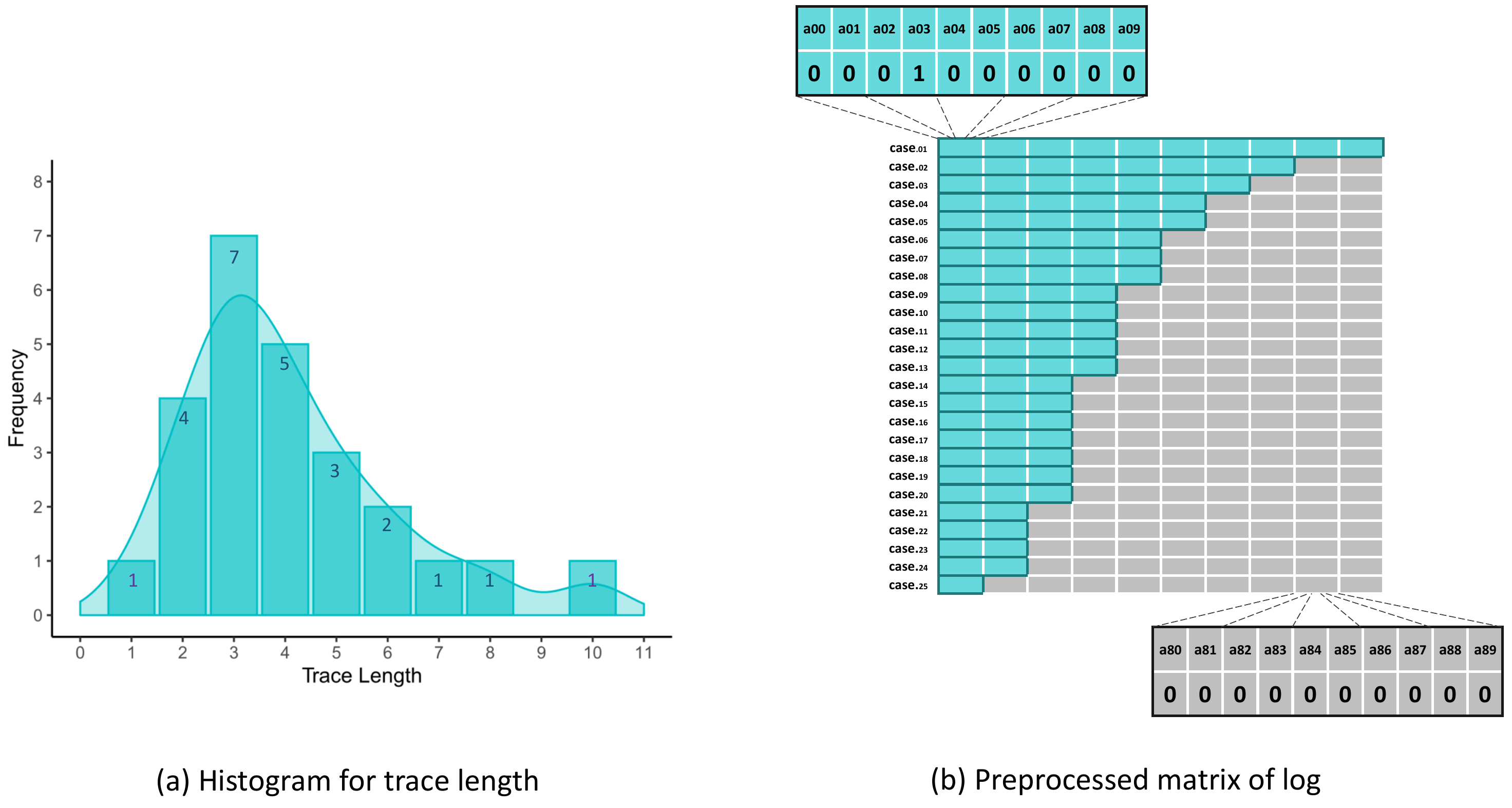}}
    \caption{Seesaw effect of zero-padding}
    \label{fig:zero-padding in matrix}
\end{figure}

This first anomaly score is likely to be biased by the zero-padded attributes in the aggregation pre-processing step. Normally, these zero-padded attributes should be treated as \texttt{null} values by any statistical method and therefore not considered in the analysis. However, this is not the case when calculating $\hat{l}(\sigma_j)$. 
The presence of 0-padded values, as shown in Figure~\ref{fig:zero-padding in matrix}, creates a a \emph{seesaw} effect that increases the leverage of longer traces and decrease the one of shorter traces. Shorter traces, in fact, are more likely to be considered similar to each other, and therefore not anomalous, because they are encoded into a higher number of zero-padded values.

In order to counter this issue, we introduce a weighting factor $w_j$ as a function of the trace length to increase/decrease the leverage $\hat{l}(\sigma_j)$  of shorter/longer traces $\sigma_j$. This weighting factor is calculated by  first normalising the trace length $N_j$ in the range $[0,1]$. This is done by applying the Z transformation to normalize $N_j$ to the average $mean[N_j]$, followed by the application of a sigmoid function. The sigmoid-based normalisation is generally used to improve the fit accuracy and decrease the computational complexity of the fitting model~\cite{klimstra2008sigmoid}: 


\begin{align}
    sig(Z_j) =\frac{1}{1+e^{-Z_j}} \text{, with } Z_j = \frac{ \{N_j - mean[N]\}}{stdev[N]}
    \label{equ:transform}
\end{align}

The weighting factor $w_j$ is then defined as:

\begin{equation}
w_j =\big[1-sig(Z_j)\big]^{c(N^{max})}
\label{eq:adjlev}
\end{equation}

The power coefficient $c(N^{max})$ is required to adjust the strength of the weighting factor for different event logs. Intuitively, if all traces in a log have similar length, then this adjustment factor should be low, approaching 1; if trace length variance is very high, then the adjustment should be higher. 

To define an appropriate value of the power coefficient $c(N^{max})$, a relation between  $N^{max}$ and the anomaly detection performance bias should be first found. However, this relation can only be estimated and not optimised because trace length has no upper bound, which would lead to a non-finite state optimisation problem. 
Therefore, we have estimated the value of $c(N^{max})$ by fitting a non-linear regression using 6 real-life event logs\footnote{These event logs belong to the ones made available by the Business Process Intelligence Challenge in 2012, 2013 and 2017} . To model a non-linear regression function, we use the values of $c(N^{max})$ that achieve the highest F1-score in anomaly detection using the 6 different logs in offline settings, i.e., considering all the traces in an event log at the same time in the observation matrix $X$. Under significance level 0.01, the non-linear equation in Eq.~\ref{equ:weighted leverage} has been fitted with two coefficient parameters $a$ and $b$ as in Table~\ref{table:regression}.

\begin{equation}
    c(N^{max}) = 
    \begin{cases}
    -2.2822 + (N^{max})^{0.3422}  &\text{   if } N^{max} > \frac{2.2822}{0.3422}  \\
    0   &\text{   otherwise}
    \end{cases}
    \label{equ:weighted leverage}
\end{equation}

\begin{table}[h]
\caption{Result of fitted non-linear regression model: $f(x)=a+x^b$ }
\vskip 0.05in
\centering
\begin{tabular}{C{2cm}C{2cm}C{3cm}C{2cm}C{2cm}}
\hline
Parameter & Estimate & Standard Error & t value & p-value \\ \hline
a         & -2.2822  & 0.3533         & -6.46   & 0.0030  \\
b         & 0.3422   & 0.0191         & 17.96   & 0.0001  \\ \hline
\end{tabular}
\label{table:regression}
\end{table}

In the end, using the estimated power coefficient $c(N^{max})$, we define the weighted \textit{leverage}-based anomaly score as:

\begin{equation}
    \hat{l}_w(\sigma_{l}) = w_j \cdot \hat{l}(\sigma_{l}).
\end{equation}

\subsection{Online anomaly detection}
\label{sec:strategy}

After having defined an anomaly score, the proposed anomaly detection method is complemented by the following four aspects: (i) grace period,  (ii) finite memory usage, (iii) update of leverage scores, and (iv) anomaly threshold setting. These are described in detail next.

\paragraph{Grace period.}

Similarly to other online anomaly detection methods in the literature~\cite{domingos2000mining,tavares2019leveraging}, for practical reasons it makes sense to begin taking decision on trace anomaly only after having received a sufficient number of events. For this purpose, we introduce the parameter Grace Period (GP), which specifies the minimum number of traces and events per trace that must be received before trace anomaly decisions can begin to be taken. In other words, the GP  prevents to run the anomaly detection model at early stages, when  a sufficient number of events has not been received yet. In this paper, the GP parameter is defined as the number of traces for which at least 2 events have been received. For example, if  GP=100, the anomaly detection starts from the first event after having received at least the first 2 events of 100 different traces.

\paragraph{Finite memory usage.}
Another condition to be satisfied by an anomaly detection method in online settings is the one of finite memory usage. In principle, events may be infinitely received as time goes by. However, handling an infinite number of events would require infinite memory, which is impossible in practical settings~\cite{tavares2019leveraging,van2019online}. Therefore, to calculate leverage using always a finite number of events received from a stream, we introduce the parameter windows size (W), defined as the number of recent cases that are considered to determine anomalies when a new event is received. More formally, at a given time $t$, let us refer to $E_{t}$ as the set of events received until $t$. Now, if the number of (possibly incomplete) traces represented by events in $E_{t}$ is more than $W$, then the earliest traces are removed from the set of traces considered to calculate the projection matrix $H$. Specifically, events of the trace whose first event is the earliest in $E_{t}$ is first removed and so on until the number of traces represented in $E_{t}$ is W.    

\begin{figure}[h]
\centering
 \resizebox{0.8\linewidth}{!}{\includegraphics{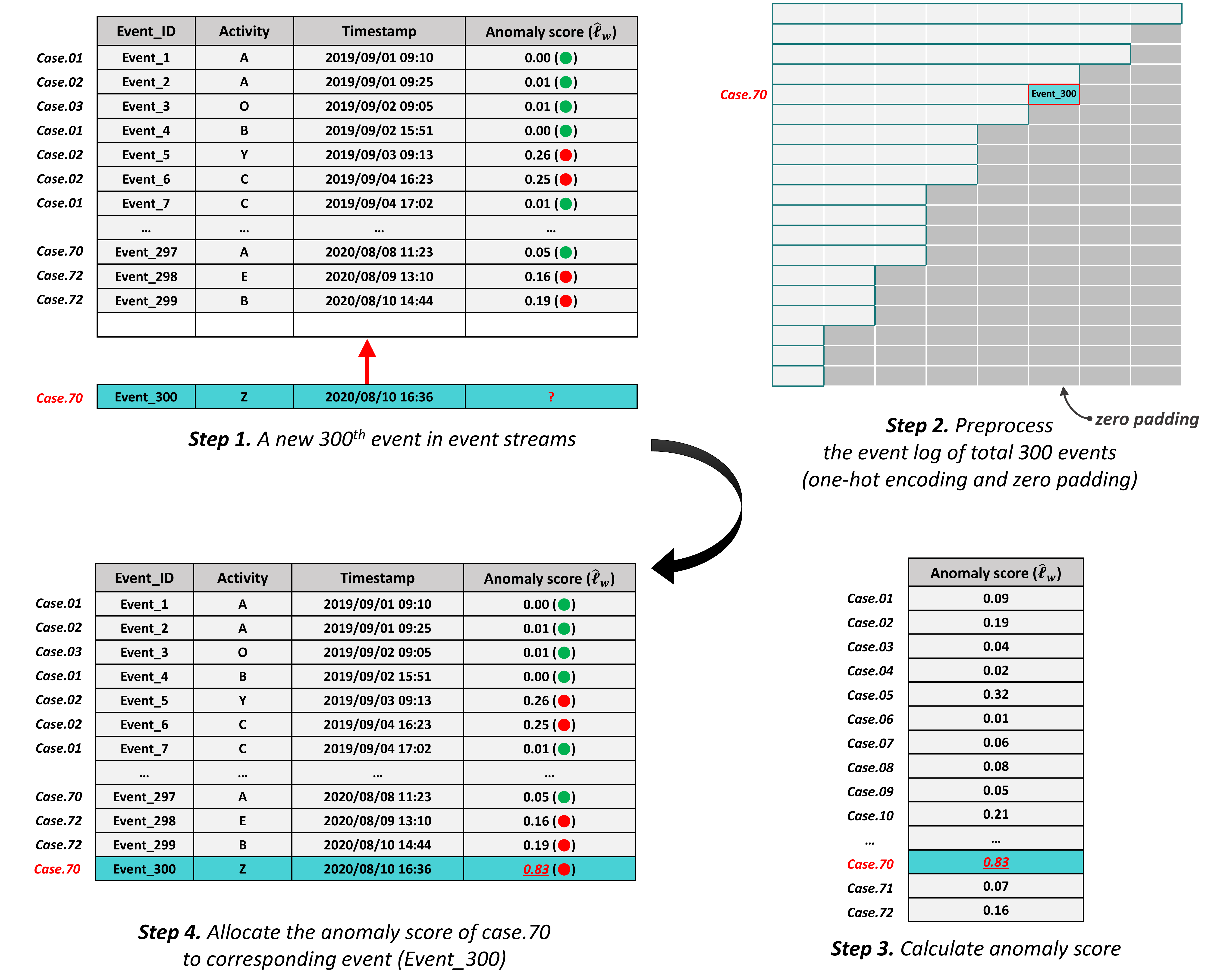}}

    \caption{Example of anomaly detection using a fixed anomaly threshold $T=0.1$}
    \label{fig:process}
    
\end{figure}

\paragraph{Update of leverage scores.}


Each time a new event is received, the leverage of the case to which this event belongs is updated (see Figure~\ref{fig:process}). More in detail, let us assume that an event $e_{i,j}$ is received at time $t$. Then, after having possibly removed some cases represented by events in $E_t$ to maintain the finite memory usage assumption, the leverage of the remaining traces $\hat{l}_w(\sigma_j)$ is calculated. The result obtained determines, based on the value of the considered anomaly detection threshold, whether the trace $\sigma_j$ is considered anomalous after the arrival of event $e_{i,j}$, or not. This procedure is replicated each time a new event is received. 


\paragraph{Anomaly threshold setting.}
The objective of anomaly detection is ultimately to determine whether traces are anomalous or not. Therefore, the problem of anomaly detection can be reduced to  a binary classification problem. Based on the anomaly score $\hat{l}_w(\sigma_{j})$, a decision should be made whether the trace $\sigma_j$ is anomalous or not. This is normally done by setting the value of an anomaly detection threshold $T$ for $\hat{l}_w(\sigma_{j})$, such that a trace is anomalous if $\hat{l}_w(\sigma_{j})>T$. In this work, we consider three constant thresholds and one variable threshold. We consider the constant values $T_{c1} = 0.1$, $T_{c2} = 0.15$, and $T_{c3} = 0.2$. These values are based on our experience in experiments  in online and offline settings, where anomalous traces tend to have an anomaly score $\hat{l}_w(\sigma_{j})$ greater than 0.1. As variable threshold, we consider the value $T_{v}=mean_{\sigma_j \subseteq E_t}[\hat{l}_w(\sigma_j)] + stdev_{\sigma_j \subseteq E_t}(\hat{l}_w(\sigma_j))$, which calculates the threshold based on the mean and standard deviation of the leverage scores of the traces considered in the observation matrix $X$. A similar principle to set the anomaly detection threshold is used by~\cite{AD10} for the timestamp anomaly detection threshold.

\section{Evaluation}
\label{sec:eval}
This section describes first the datasets that we used for evaluating the proposed online anomaly detection framework. Then, we present the evaluation metrics and experiment settings and, finally, we discuss the performance and computational cost of the proposed framework.


We consider two artificial logs used by~\cite{AD11} and one real-life log publicly available. The artificial logs are generated by simulating 2 process models (Small and Medium in~\cite{AD11}) using the PLG2 tool. Regarding the real log, we consider the \textit{Helpdesk} event log, which contains events logged by a ticketing management system of the help desk of an Italian software company. These logs have been chosen because they have been considered by previous work in anomaly detection and they are also sufficiently small to control the running time of experiments. Descriptive statistics of these logs are reported in Table~\ref{table:data}. 

\begin{table}[t]
\caption{Descriptive statistics of event logs (Log statistics are counted after injecting anomalies)}
\vskip 0.03in
\centering
\begin{scriptsize}
\begin{tabular}{C{2.5cm}|C{2.1cm}|L{4.8cm}|R{1.7cm}}
\hline
Type                             & Data                           & \qquad\qquad\quad Statistics                     & Value \\ \hline
\multirow{10}{*}{Artificial log} & \multirow{5}{*}{Small}         & \,-Number of cases                & 5,000      \\
                                 &                                & \,-Number of events               & 44,811     \\
                                 &                                & \,-Number of activities           & 20        \\
                                 &                                & \,-Average \# of cases per day  & 5,000      \\
                                 &                                & \,-Average \# of events per day & 44,811     \\ \cline{2-4}
                                 & \multirow{5}{*}{Medium}        & \,-Number of cases                & 5,000      \\
                                 &                                & \,-Number of events               & 29,683     \\
                                 &                                & \,-Number of activities           & 32        \\
                                 &                                & \,-Average \# of cases per day  & 5,000      \\
                                 &                                & \,-Average \# of events per day & 29,683     \\ \hline
\multirow{5}{*}{Real log}        & \multirow{5}{*}{Helpdesk} & \,-Number of cases                &  3,804      \\
                                 &                                & \,-Number of events               & 13,901     \\
                                 &                                & \,-Number of activities           & 9        \\
                                 &                                & \,-Average \# of cases per day  & 10.94      \\
                                 &                                & \,-Average \# of events per day & 18.06     \\ \hline
\end{tabular}
\end{scriptsize}
\vskip -0.07in
\label{table:data}
\end{table}

Evaluating an unsupervised approach of anomaly detection like the one that we propose requires event logs with labelled traces (normal v. anomalous), which are generally unavailable in practice. Therefore, a common practice in this research field  is to inject anomalies using different types of anomaly patterns in event log and creating labels during the anomaly injection process~\cite{AD2,AD9,AD12,AD10,AD11}. We consider the 5 anomaly patterns \textit{Skip}, \textit{Insert}, \textit{Early}, \textit{Late}, and \textit{Rework} as defined in~\cite{AD11}: in $Skip$, a sequence of events is skipped in some cases; in $Insert$, one or more events are generated in random positions within existing traces; in $Early$/$Late$, timestamps of events are manipulated such that a sequence of events is moved earlier/later in a trace; in $Rework$, a sequence of events is repeated after its occurrence. Anomalies are randomly injected in an event log until 10\% of the traces in the log have become anomalous. 
As far as performance measures are concerned, we consider the typical measures for classification problems obtained from the confusion matrix, i.e., precision, recall and F1-score.  
The datasets used in this paper and the code to reproduce the experiments discussed next are available at \url{https://github.com/jonghyeonk/OnlineAnomalyDetection}.


We set the GP to 1,000 cases and consider 3 values of window size W, i.e., $W \in [1000, 2000, 3000]$. A larger value of GP and W is likely to lead to better and more stable performance, while also implying a higher computational cost. 
The experiments are implemented in R on an Intel i7 Linux machine using a single CPU and 5GB memory limit.


\begin{table}[t]
\caption{Performance of online anomaly detection (average from the start of the stream and calculated only on first/last 100 events)}
\vskip 0.03in
\centering
\resizebox{0.8\textwidth}{!}{%
\begin{tabular}{C{1.5cm}|C{2cm}|C{1.9cm}|L{1.6cm}|C{1.0cm}|C{1.0cm}|C{1.0cm}|C{1.0cm}|C{1.0cm}|C{1.0cm}}
\hline
\multirow{2}{*}{Data}     & \multirow{2}{*}{Window size} & \multirow{2}{*}{\begin{tabular}[c]{@{}c@{}}Time cost\\      (average sec)\end{tabular}} & \multirow{2}{*}{Metric} & \multicolumn{6}{c}{Threshold}                                              \\ \cline{5-10}
                          &                              &                                                                                         &                         & $T_{c1}$ & $T_{c2}$ & $T_{c3}$ & $T_{v}$ & $T_{v}^{F:100}$ & $T_{v}^{L:100}$ \\ \hline
\multirow{9}{*}{Small}    & \multirow{3}{*}{1,000}       & \multirow{3}{*}{1.09}                                                                   & Precision               & 0.22     & 0.22     & 0.22     & 0.21    & 0.06           & 1.00              \\ \cline{4-10} 
                          &                              &                                                                                         & Recall                  & 0.63     & 0.62     & 0.59     & 0.62    & 1.00              & 0.63           \\ \cline{4-10} 
                          &                              &                                                                                         & F1-score                & 0.33     & 0.32     & 0.32     & 0.31    & 0.11           & 0.77           \\ \cline{2-10} 
                          & \multirow{3}{*}{2,000}       & \multirow{3}{*}{1.07}                                                                   & Precision               & 0.26     & 0.25     & 0.25     & 0.25    & 0.06           & 1.00              \\ \cline{4-10} 
                          &                              &                                                                                         & Recall                  & 0.61     & 0.60      & 0.57     & 0.61    & 1.00              & 0.63           \\ \cline{4-10} 
                          &                              &                                                                                         & F1-score                & 0.36     & 0.36     & 0.35     & 0.35    & 0.11           & 0.77           \\ \cline{2-10} 
                          & \multirow{3}{*}{3,000}       & \multirow{3}{*}{1.23}                                                                   & Precision               & 0.75     & 0.79     & 0.82     & 0.75    & 0.20            & 1.00              \\ \cline{4-10} 
                          &                              &                                                                                         & Recall                  & 0.57     & 0.55     & 0.53     & 0.57    & 0.17           & 0.63           \\ \cline{4-10} 
                          &                              &                                                                                         & F1-score                & 0.65     & 0.65     & 0.64     & 0.65    & 0.18           & 0.77           \\ \hline
\multirow{9}{*}{Medium}   & \multirow{3}{*}{1,000}       & \multirow{3}{*}{1.42}                                                                   & Precision               & 0.16     & 0.17     & 0.17     & 0.15    & 0.50            & 1.00              \\ \cline{4-10} 
                          &                              &                                                                                         & Recall                  & 0.72     & 0.71     & 0.71     & 0.73    & 0.29           & 0.50            \\ \cline{4-10} 
                          &                              &                                                                                         & F1-score                & 0.26     & 0.27     & 0.27     & 0.25    & 0.36           & 0.67           \\ \cline{2-10} 
                          & \multirow{3}{*}{2,000}       & \multirow{3}{*}{1.46}                                                                   & Precision               & 0.37     & 0.45     & 0.52     & 0.26    & 0.50            & 1.00              \\ \cline{4-10} 
                          &                              &                                                                                         & Recall                  & 0.65     & 0.64     & 0.63     & 0.68    & 0.29           & 0.50            \\ \cline{4-10} 
                          &                              &                                                                                         & F1-score                & 0.47     & 0.53     & 0.57     & 0.38    & 0.36           & 0.67           \\ \cline{2-10} 
                          & \multirow{3}{*}{3,000}       & \multirow{3}{*}{2.06}                                                                   & Precision               & 0.28     & 0.30      & 0.31     & 0.25    & 0.50            & 1.00              \\ \cline{4-10} 
                          &                              &                                                                                         & Recall                  & 0.67     & 0.67     & 0.66     & 0.67    & 0.29           & 0.50            \\ \cline{4-10} 
                          &                              &                                                                                         & F1-score                & 0.39     & 0.41     & 0.42     & 0.37    & 0.36           & 0.67           \\ \hline
\multirow{9}{*}{Helpdesk} & \multirow{3}{*}{1,000}       & \multirow{3}{*}{0.34}                                                                   & Precision               & 0.06     & 0.06     & 0.06     & 0.06    & 0.00              & 0.50            \\ \cline{4-10} 
                          &                              &                                                                                         & Recall                  & 0.99     & 0.98     & 0.96     & 1.00       & 0.00              & 0.96           \\ \cline{4-10} 
                          &                              &                                                                                         & F1-score                & 0.12     & 0.12     & 0.12     & 0.12    & 0.00              & 0.66           \\ \cline{2-10} 
                          & \multirow{3}{*}{2,000}       & \multirow{3}{*}{0.37}                                                                   & Precision               & 0.08     & 0.08     & 0.09     & 0.08    & 0.00              & 0.51           \\ \cline{4-10} 
                          &                              &                                                                                         & Recall                  & 0.80      & 0.74     & 0.68     & 0.77    & 0.00              & 0.96           \\ \cline{4-10} 
                          &                              &                                                                                         & F1-score                & 0.15     & 0.15     & 0.15     & 0.14    & 0.00              & 0.67           \\ \cline{2-10} 
                          & \multirow{3}{*}{3,000}       & \multirow{3}{*}{0.39}                                                                   & Precision               & 0.09     & 0.11     & 0.13     & 0.10     & 0.00              & 0.61           \\ \cline{4-10} 
                          &                              &                                                                                         & Recall                  & 0.59     & 0.49     & 0.39     & 0.60     & 0.00              & 0.93           \\ \cline{4-10} 
                          &                              &                                                                                         & F1-score                & 0.16     & 0.19     & 0.20      & 0.17    & 0.00              & 0.74           \\ \hline
\end{tabular}
}
\label{table:performance}
\end{table}

Table~\ref{table:performance} shows the performance of anomaly detection in event streams for different anomaly detection threshold values and different values of W. Note that the 4 columns $T_{c1}$ to $T_{v}$ report average performance measures calculated from the start of the streaming (after the GP condition has been reached).  It can be noticed that the three constant thresholds show on average better performance than the variable threshold $T_{v}$. 
To better observe a trend of performance improvement as more events are received, the last two columns $T_v^{F:100}$ and $T_v^{L:100}$ show the average performance values calculated on the first 100 events received (after the GP condition has been met) and last 100 events received, respectively. The result shows a clear tendency of increasing performance. The performance is low at the initial stage, and it increases remarkably for the last 100 events received. Particularly in the case of the $Helpdesk$ log, the proposed framework could not detect any anomalous cases in the first 100 events, while the performance clearly increases for the last 100 events. 
It should be noted that, as the number of events received increases, the performance of the proposed framework is likely to converge to the one showed by the average on the last 100 events. Regarding the window size W, the average time cost increases with the value of W.  A larger window size also leads to better anomaly detection performance. 

\begin{figure}[t]
    \centering
    \makebox[0pt]{%
    \includegraphics[width=0.98\textwidth]{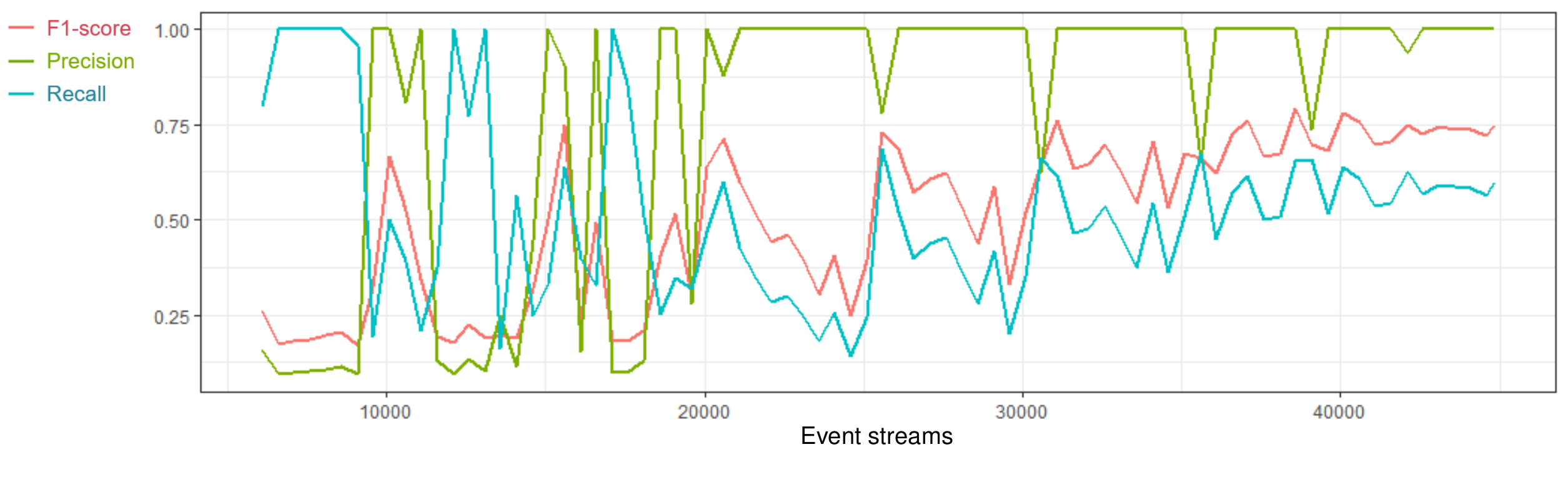}}
    \vskip -0.03in
    \caption{Performance values as number of events received increases (Small log is applied, using $T_{v}$, W=3000)}
    \label{fig:trend}
    \vskip -0.07in
\end{figure}

As an example, Figure~\ref{fig:trend} breaks down the performance of the proposed framework along time, counted as the number of events received, in the case of the Small event log with W=3000. It can be noted that the performance oscillates wildly until 20,000 events are received. After that, the performance tends to stabilise and, while the precision remains high, the recall also begins increasing more regularly. After 30,000 events are received, all the performance metrics appear to have become stable.

\section{Conclusions}
\label{sec:conclu}

This paper has presented an approach to detect trace level anomalies in business process event streams using an anomaly score based on statistical leverage. A preliminary evaluation on artificial and real event logs also has been presented. 
The results obtained in this paper are important to determine the future work in this line of research. 

First, the performance in the case of the $Helpdesk$ log highlights an issue with anomaly threshold setting. The constant values chosen for the experiment (between 0.1 and 0.2) appear to be too low for this event log, which  results in low precision and F1-score and only high recall. This points to the need for developing an advanced variable threshold that can adapt to the characteristics of different event logs.  


Another limitation of the proposed approach, which also impacts the performance, is the fact that it does not distinguish between incomplete and completed traces when calculating the anomaly score. Therefore, many traces may be considered anomalous because they are incomplete, even though they will turn into normal at some point in the future as more events are received. A possible strategy to prevent this is to organise the events received into batches  by different prefix length before calculating the anomaly score. This is inspired by~\cite{van2019online} that, in the case of online compliance checking, addresses the issue of trace incompleteness using prefix-alignment. 

Finally, considering word-embedding instead of one-hot encoding and zero-padding during pre-processing may be likely to reduce the size of the observation matrix $X$ and, therefore, speed up the calculation of anomaly scores.

\bibliographystyle{abbrv}
\bibliography{biblio}

\end{document}